\documentclass[conference]{IEEEtran}

\usepackage[utf8]{inputenc}
\usepackage{circuitikz}
\usepackage{subcaption}
\usepackage{cite}
\usepackage{amsmath,amssymb,amsfonts}

\usepackage{graphicx}
\usepackage{textcomp}
\usepackage{xcolor}
\usepackage{enumitem}
\usepackage{url}
\usepackage{circuitikz}
\usepackage{stfloats} % Add this package
\usepackage[export]{adjustbox}%multi figure
\usepackage{lipsum}
\usepackage{mwe}
\usepackage{todonotes}
\usepackage{booktabs}

\usepackage{algpseudocode}
\usepackage{algorithm}

\def\BibTeX{{\rm B\kern-.05em{\sc i\kern-.025em b}\kern-.08em
    T\kern-.1667em\lower.7ex\hbox{E}\kern-.125emX}}
    
\begin{document}

\title{Event-Driven Digital-Time-Domain Inference Architectures for Tsetlin Machines\\

\author{
Tian Lan$^\dagger$, Rishad Shafik$^\dagger$, Alex Yakovlev$^\dagger$\\
$^\dagger$Microsystems Research Group, Newcastle University, Newcastle upon Tyne, UK\\
\textrm{\{t.lan3, rishad.shafik, alex.yakovlev\}@newcastle.ac.uk}
}

}

\maketitle

\begin{abstract}

Machine learning fits model parameters to approximate input-output mappings, predicting unknown samples. However, these models often require extensive arithmetic computations during inference, increasing latency and power consumption. This paper proposes a digital-time-domain computing approach for Tsetlin machine (TM) inference process to address these challenges. This approach leverages a delay accumulation mechanism to mitigate the costly arithmetic sums of classes and employs a Winner-Takes-All scheme to replace conventional magnitude comparators. Specifically, a Hamming distance-driven time-domain scheme is implemented for multi-class TMs. Furthermore, differential delay paths, combined with a leading-ones-detector logarithmic delay compression digital-time-domain scheme, are utilised for the coalesced TMs, accommodating both binary-signed and exponential-scale delay accumulation issues. Compared to the functionally equivalent, post-implementation digital TM architecture baseline, the proposed architecture demonstrates orders-of-magnitude improvements in energy efficiency and throughput.

\end{abstract}


\begingroup
\renewcommand\thefootnote{}\footnotetext{
This work was supported by EPSRC EP/X036006/1 Scalability Oriented Novel Network of Event Triggered Systems (SONNETS) project and by EPSRC EP/X039943/1 UKRI-RCN: Exploiting the dynamics of self-timed machine learning hardware (ESTEEM) project.
}
\addtocounter{footnote}{0}
\endgroup

\section{Introduction}
Edge computing has emerged as a key enabler for real-time machine learning (ML) applications, driven by characteristics of low‐latency responses, reduced bandwidth demands, and enhanced security \cite{7807196} \cite{10.1145/2627369.2631634}. Applications like drone collision avoidance, autonomous driving, and augmented reality \cite{10.1109/SECON48991.2020.9158429} rely critically on rapid inference and low communication overhead, making ML algorithms deployed at the edge particularly appealing. However, edge devices are often deployed far from gateways and frequently operate within limited energy budgets. These constraints impose stringent energy management on ML algorithms implemented at the edge.

In edge ML energy management, two pressing contradictions require resolutions. \textbf{First}, synchronous systems are controlled by a global clock whose frequency must exceed the incoming‐event rate and cannot be adapted dynamically as that rate fluctuates, resulting in significant energy waste. As a resolution, event-driven asynchronous ML architectures \cite{10239577} \cite{10.1109/TVLSI.2024.3464870} enable processing only when events occur, eliminating clock power during idle intervals; this sparse computing consequently achieves lower average energy consumption and improved performance.

\textbf{Second}, ML algorithms are inherently intensive in arithmetic operations. Some examples include: (i) the evaluation of inner products using convolutional kernels in convolutional neural networks (CNNs); (ii) dense matrix multiply-accumulate (MAC) operations in fully connected layers of deep neural networks (DNNs); (iii) the accumulation of membrane potentials in spiking neural networks (SNNs); and (iv) various nonlinear transforms such as ReLU, softmax, argmax, and their variants that convert weighted sums into class-conditional probabilities or discrete decisions. Implementing these essential operations in the digital domain requires many arithmetic logic units (ALUs), MAC arrays, and magnitude comparators (MCs). Each logic occurs through discrete voltage transitions, with every transition charging or discharging the effective capacitance of CMOS nodes. This switching activity generates significant dynamic power, resulting in substantial energy dissipation per inference, especially at the edge \cite{9661725}. As a weak-capacitance approach, transferring the extensive arithmetic operations to the time domain shows considerable promise. Recent studies \cite{9987520} \cite{10318175} have demonstrated that the MAC operations of NNs can be represented as pulse-delay accumulation. This method greatly reduces the charging and discharging of CMOS nodes, providing a practical route toward ultra-low-power solutions for edge devices.

Multi-class Tsetlin Machines (TMs) and variants such as the Coalesced TM (CoTM) represent a promising category of ML algorithms based on propositional logic \cite{granmo2021tsetlinmachinegame} \cite{glimsdal2021coalescedmultioutputtsetlinmachines}. They offer several advantages for edge-oriented AI hardware: (i) \textbf{Hardware Affinity}: TM inference is based on Boolean operations, eliminating the need for convolutional layers and large-scale matrix multiplications. This allows for straightforward mapping of field-programmable gate array (FPGA) or application-specific integrated circuit (ASIC) architectures. (ii) \textbf{Simple Inference}: The inference processes in TMs are conceptually simple and computationally lightweight, making them well-suited for satisfying real-time and low-power requirements. (iii) \textbf{Full Interpretability}: Every prediction made by a TM can be traced back to explicit Boolean clause combinations. This ensures that decision pathways are transparent and understandable to humans, rather than relying on the opaque internal representations of DNNs. Collectively, these properties position TMs as an energy-efficient, fast, and explainable alternative to NNs for edge ML tasks.

Deploying TM's inference process at the edge still needs to address the previously identified challenges of eliminating the global clock and alleviating intensive arithmetic operations. For multi-class TMs, it has been demonstrated that both asynchronous (four-phase handshake) dual-rail architectures \cite{9474126} and time-domain strategies \cite{10666312} are able to achieve energy efficiency. This paper presents design paradigms for multi-class TM and CoTM using asynchronous bundled data (BD) and hybrid digital-time-domain methods. Energy efficiency and throughput are quantitatively evaluated against functionally equivalent, post-implementation synchronous and asynchronous hardware, demonstrating the advantages of the proposed architecture.

\textbf{Contributions}:
\begin{itemize}[leftmargin=*]
\item Novel hybrid digital-time-domain computing paradigms are proposed that integrate seamlessly with event-driven asynchronous logic. 
\item The proposed architecture realises an ultra-low-power, high-throughput inference process.
\item Functionally equivalent synchronous and asynchronous digital TM inference circuits are implemented, providing a direct baseline for the proposed approach.
\end{itemize}

\section{Methodology}

\begin{figure}[h]
\setlength\abovecaptionskip{0.1\baselineskip}
\setlength\belowcaptionskip{0.1\baselineskip}
   \centering
   \includegraphics[width=0.49\textwidth]{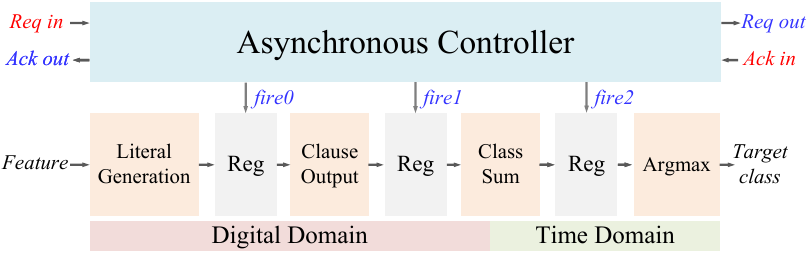}
    \vspace{0mm}
    \caption{\textcolor{black}{Block diagram of the proposed architecture.}}
    \label{figure:bd}
\end{figure}

Fig. \ref{figure:bd} depicts the block diagram of the proposed architecture. At the top hierarchical level, an asynchronous controller synchronises the downstream functional modules by forwarding a $Request$ signal from $Req\_in$ to $Req\_out$. The $Acknowledge$ path runs from $Ack\_in$ to $Ack\_out$, providing reverse feedback to complete each transaction. The functional modules are designed as a hybrid data path. Specifically, literal generation and clause output are carried out in the digital domain. The class sum and argmax functions are converted to the time domain.

\vspace{0mm}
\vspace{-5mm}

\subsection{Asynchronous Controller}

\begin{figure}[]
\setlength\abovecaptionskip{0.1\baselineskip}
\setlength\belowcaptionskip{0.1\baselineskip}
   \centering
   \includegraphics[width=0.5\textwidth]{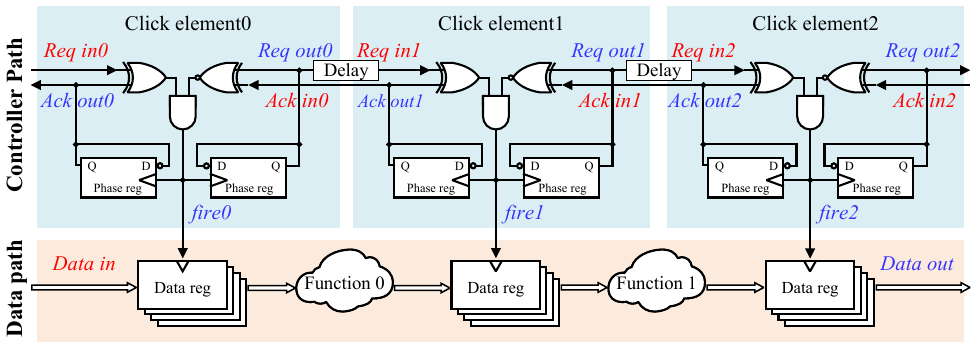}
    \vspace{0mm}
    \caption{\textcolor{black}{Click element-based asynchronous pipeline.}}
    \vspace{0mm}
    \label{figure:pipeline}
\end{figure}

The asynchronous controller illustrated in Fig.~\ref{figure:pipeline} is constructed as a three-stage Click element. The two-phase handshake protocol achieves \textbf{Elastic throughput}: Computation proceeds only when data is available, eliminating idle switching and the global clock to handle worst-case path delays. \textbf{Gate-level implementation}: the circuit employs combinatorial logic and flip-flops, enabling direct synthesis with standard CMOS workflows and ensuring compatibility with electronic design automation (EDA) toolkits ~\cite{10239577} \cite{10.1109/TVLSI.2024.3464870} \cite{5476997}. The pseudocode for a click element stage is listed in Algorithm \ref{alg:click}.

\vspace{-5mm}

\begin{algorithm}[H]
  \caption{Click Element-based Asynchronous Controller}
  \label{alg:click}
  \begin{algorithmic}[1]
    \Function{ClickElement}{}
    \State \textbf{Iutput} $rst,req\_in, ack\_in$
      \State \textbf{Output} $req\_out,\; ack\_out,\; fire$
      \State \textbf{Logic}  $phase\_in,\; phase\_out$
      \If{$rst$}
        \State $phase\_in \gets 0$\; ;\; $phase\_out \gets 0$
      \Else
        \State $fire \gets (req\_in \oplus phase\_in)\wedge
                          \neg(ack\_in \oplus phase\_out)$
        \If{$fire\uparrow$}
          \State $phase\_in  \gets \neg phase\_in$
          \State $phase\_out \gets \neg phase\_out$
        \EndIf
      \EndIf
      \State $req\_out \gets phase\_in$
      \State $ack\_out \gets phase\_out$
      \State \Return $req\_out, ack\_out, fire$
    \EndFunction
  \end{algorithmic}
\end{algorithm}
\vspace{-5mm}

\subsection{Clause Evaluation}
The literal generation and clause output stages are combined
into the clause evaluation functional module. The descriptive
hardware pseudocode is listed in Algorithm \ref{alg:clauseE}.

\vspace{0mm}
\begin{algorithm}[H]
\caption{Clause Evaluation}
\label{alg:clauseE}
\begin{algorithmic}[1]
\Function{ClauseEvaluation}{}
  \State \textbf{Input} $rst$, $fire0$, $feature$  $\in \{0,1\}^{\text{num\_feature}}$
  \State \textbf{Output} $clause\_vector$ $\in \{0,1\}^{\text{num\_clause}}$
  \State \textbf{Logic} $literal$ $\in \{0,1\}^{\text 2\times{num\_feature}}$
  \If{$rst$}
    \State $clause\_vector$ \(\gets 0\)
  \ElsIf{$fire0\uparrow$}
    \For{\(i \gets 0\) to \(num\_feature - 1\)}
      \State $literal[2i]$ \(\gets\) $feature[i]$
      \State $literal[2i + 1]$ \(\gets\) \(\neg\)$feature[i]$
    \EndFor

    \For{\(j \gets 0\) to \(num\_clause - 1\)}
      \State $clause\_vector[j]$ \(\gets \bigwedge (\,literal\,\lor\, TA\_states)\)
    \EndFor
  \EndIf

  \State \Return $clause\_vector$
\EndFunction
\end{algorithmic}
\end{algorithm}

\subsection{Classification}

\begin{figure*}[t]
    \centering
    \includegraphics[width =1\textwidth]{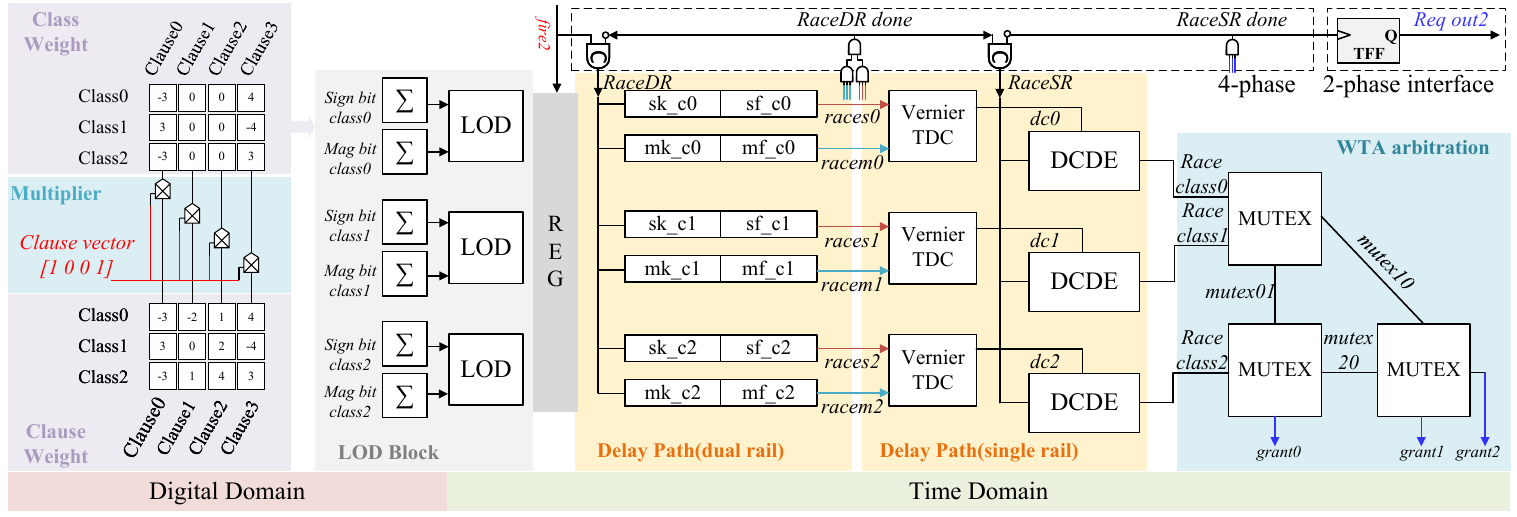}
    \vspace{0mm}
    \caption{\textcolor{black}{The hybrid digital-time domain architecture for CoTM classification.}}
    \label{figure:cotmc}
\end{figure*}

Classification is a significant functional module in the TM inference process, which aggregate the weighted contributions of active clauses to reach the final classification. Using hardware description languages (listed in Algorithm \ref{alg:class}) to build this module is intuitive and convenient. However, conducting extensive arithmetic operations purely in the digital domain requires numerous ALUs, which leads to significant energy consumption \cite{10666312}. Compared to the energy used for memory accesses, the energy overhead from arithmetic operations can be more effectively reduced through architectural optimisation. The classification function for multi-class TM inference can be defined as follows:

\begin{equation}
\hat{y} =  \operatorname*{argmax}_{i=1,\dots,m} \left( \sum_{j=1}^{\frac{n}{2}} C_j^{1,i}(X) - \sum_{j=1}^{\frac{n}{2}} C_j^{0,i}(X) \right)
\label{eq:classification}
\end{equation}

\begin{algorithm}
\caption{Classification}
\label{alg:class}
\begin{algorithmic}[1]
\Function{Classification}{}
    \State \textbf{Input} $fire1, fire2,$ $clause\_vector$ $\in \{0,1\}^{\text{num\_clause}}$
    \State \textbf{Output} $target\_class$
    \State \textbf{Logic} $sum\_weight$ $,class\_sum$ $\in \mathbb{Z}^{\text{num\_class}}$
    \If{$rst$}
        \State $sum\_weight$ $\gets$ 0
    \ElsIf{$fire1$ $\uparrow$}
        \State $sum\_weight$ $\gets$ $clause\_vector$ $\wedge$ $clause\_weight$
    \EndIf

    \If{$fire2$ $\uparrow$}
        \For{$i \gets 0$ \textbf{to} \text{$num\_class$} $- 1$}
            \State $class\_sum[i]$ $\gets$ \text{Sum}($sum\_weight[i]$)
        \EndFor
        \State $target\_class$ $\gets$ \text{Argmax}($class\_sum$)
    \EndIf

    \State \Return $target\_class$
\EndFunction
\end{algorithmic}
\end{algorithm}

This function can be regarded as comparing Hamming distances among different classes, where contributions from ones in positive clauses and zeros in negative clauses are considered equivalent, as are contributions from zeros in positive clauses and ones in negative clauses. A detailed time-domain classification architecture has already been discussed in \cite{10666312}. This work further develops a streamlined interface that seamlessly integrates the two-phase BD controller with the four-phase quasi-delay-insensitive (QDI) classification module, which will later be introduced in the CoTM part.

\begin{equation}
\hat{y} = \operatorname*{argmax}_{i=1,\dots,m} \left(\sum_{j=1}^{n} W_j^{i}\cdot \mathcal{C}_j^{\,i}(X)\right)
\label{eq:cotmc}
\end{equation}

In the CoTM inference process, the classification function (Eq. \ref{eq:cotmc}) selects the target class with the highest binary MAC value. The CoTM does not pre-assign positive or negative clauses; instead, any clause can either support (with a positive integer weight) or oppose (with a negative integer weight) any class, greatly enhancing expressiveness. However, this flexibility introduces two significant challenges in designing time-domain hardware: (i) \textbf{Signed class sum addition:} The unidirectionality of time encoding makes it challenging to represent signed weight summation directly. (ii)\textbf{Exponential path delay growth:} As the number of clauses increases, the path delay for a class grows exponentially, which inflates area requirements and decreases robustness to process–voltage–temperature (PVT) variations. 

Applied strategies: \textbf{(i) Encode class sum into differential delay path}: Digital logic pre-calculates all clause \emph{sign} contributions into \( S \) and all \emph{magnitude} contributions into \( M \). Then, launches two independent pulses (\textit{race}$_S$ and \textit{race}$_M$). The interval difference encodes the final signed class sum. \textbf{(ii)Leading-ones-detector (LOD) delay compression}: Coarse delays \((s^k, m^k)\) logarithmically segment the delay range, while a \( e \)-bit normalised fine delay \((s^f, m^f)\) is used for high-resolution tuning \cite{823449}.

The proposed CoTM classification implementation (illustrated in Fig.~\ref{figure:cotmc}) leverages a digital-time-domain architecture. The main functional modules include: a binary multiplication matrix, an LOD block, a delay accumulation module, a Winner-Takes-All (WTA) arbitration system, a race control unit, and a four-to-two phase interface for asynchronous communication. \textbf{Specifically}:

\subsubsection{\textbf{Binary multiplication matrix}}
This module selects the weights of activated clauses based on the clause evaluation results. The chosen clause weights are then used to compute the subsequent class sum. This can be implemented as a multiplier (MUX) array.

\subsubsection{\textbf{LOD Coarse Fine Delay Extraction}}

As a front-end circuit to the delay accumulation module, the LOD block mitigates the exponential growth of delay paths caused by the increasing bit width of the class sum. The pseudocode hardware description is listed in Algorithm~\ref {alg:lod}.

\begin{algorithm}
\caption{LOD coarse fine delay extraction}
\label{alg:lod}
\begin{algorithmic}[1]
\Function{LOD}{}
  \State \textbf{Input} $\textit{SumValue}$, $e$
  \State \textbf{Output} $k,f$
  \State \textbf{Logic} $BitWidth,mask$
  \For{$i = BitWidth-1$ \textbf{downto} $0$}
      \If{$\textit{SumValue}[i]$}
         \State $k \gets i$
         \State \textbf{break}
      \EndIf
  \EndFor
  \State $mask \gets (1 \ll k) - 1$
  \State $f \gets SumValue \,\&\, mask$
  \If{$k \ge e$}
     \State $f \gets f \gg (k - e)$
  \Else
     \State $f \gets f \ll (e - k)$
  \EndIf
  \State \Return $k,f$
\EndFunction
\end{algorithmic}
\end{algorithm}

With the values of \( S \) or \( M \) (\(\textit{SumValue}\)), the LOD first executes a leading-ones detection to determine the coarse delay index \( k \). This process maps the pulse to a logarithmically encoded delay segment. Next, the residual bits below \( k \) are extracted and normalised to a \( e \)-bit resolution, which is then returned as the fine delay \( f \). By decomposing the overall path delay into pairs $(k,f)$, the LOD compresses an exponential path space to logarithmic complexity while maintaining timing resolution.

\subsubsection{\textbf{Delay accumulation}}

\begin{figure}[h]
\setlength\abovecaptionskip{0.1\baselineskip}
\setlength\belowcaptionskip{0.1\baselineskip}
   \centering
   \includegraphics[width=0.5\textwidth]{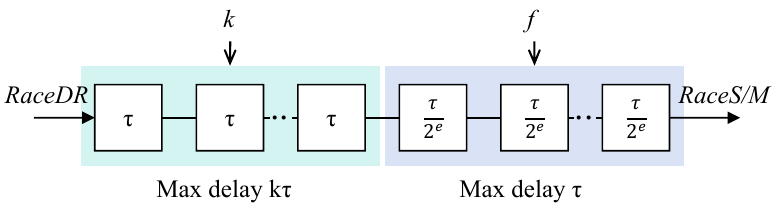}
    \vspace{0mm}
    \caption{\textcolor{black}{Differential delay path.}}
    \label{figure:cfd}
\end{figure}

The delay accumulation module comprises a differential delay path, a Vernier time-to-digital converter (TDC) \cite{823449}, and a single-rail (SR) delay path. The differential delay path (Fig.~\ref{figure:cfd}) is controlled by the coarse fine parameter $(k,f)$ generated by the LOD block, where the coarse unit delay is $\tau$ and the maximum coarse delay is $k\,\tau$. The fine unit delay is $\tau/2^{e}$, ensuring that the maximum span of fine delay remains \(\tau\) while permitting a resolution of \(e\). With the simultaneous launch event \textit{raceDR}, the arrival instants of the two rails (\textit{raceS} and \textit{raceM}) encode the sign and magnitude of the class sum. A Vernier TDC digitises these race pulses to produce a compact delay code $dc$. Subsequently, $dc$ configures a digitally-controlled delay element (DCDE) to realise the final SR delay path; because of the delay compression, this stage requires only a short length. Typical DCDE implementations may employ digital multiplexed segments \cite{6630119,10666312}, or analogue structures such as current-starved inverters \cite{375961} and shunt-capacitor inverters \cite{doi:10.1049/iet-cdt:20060111}.

\vspace{0mm}
\vspace{0mm}

\subsubsection{\textbf{WTA arbitration}}

The WTA arbitration module monitors the rising edge of the concurrent $m$ race signals, denoted as $\textit{RaceClass}[m\!-\!1{:}0]$. It grants the first-arriving pulse, selecting the class index with the highest arithmetic value. Since the output $\textit{grant}[m\!-\!1{:}0]$ is a one-hot vector, the WTA also acts as the terminal of the time-domain path, interfacing directly with the subsequent digital-domain ports. Two WTA implementations are supported:

\begin{enumerate}[label=(\roman*)]
  \item \textbf{Tree-Based Arbiter (TBA)} \cite{10666312}: The TBA uses a QDI hierarchy that propagates local winner signals forward until a global winner is recognised. For a system with \( m \) competing classes, this architecture requires \(\lceil\log_{2} m\rceil\) arbitration layers arranged in a binary tree, along with a total of \( m-1 \) identical TBA cells.
  \item \textbf{Mesh-Like Arbiter} \cite{Golubcovs2011MultiresourceAT}: This topology is an all-pair cyclic-comparison network; the final winner emerges after $m-1$ stages and is implemented using $\dfrac{m(m-1)}{2}$ mutual exclusion (Mutex) cells.
\end{enumerate}

\vspace{0mm}
\begin{figure}[h]
\setlength\abovecaptionskip{0.1\baselineskip}
\setlength\belowcaptionskip{0.1\baselineskip}
   \centering
   \includegraphics[width=0.32\textwidth]{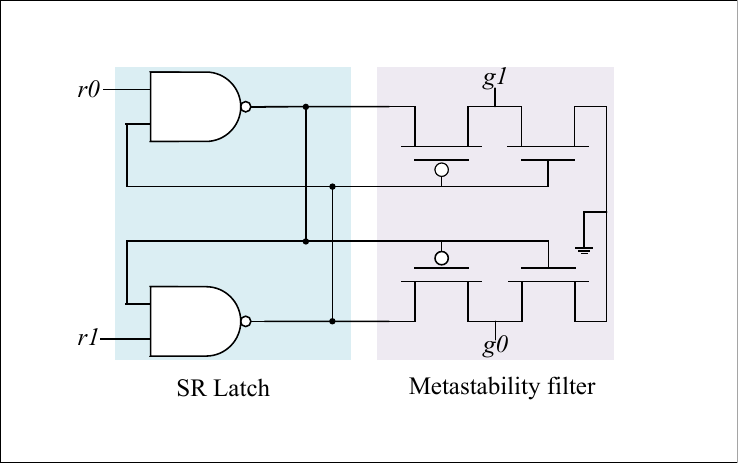}
    \vspace{0mm}
    \caption{\textcolor{black}{Gate-level netlist of a Mutex}}
    \label{figure:mutex}
\end{figure}
\vspace{0mm}

Figure~\ref{figure:mutex} shows the gate‐level netlist of the Mutex \cite{8097382}. This component consists of a Set-Reset latch formed by two cross-coupled NAND gates, which detect the rising edge of competing signals. Specifically, when a faster input transitions from low to high, the corresponding NAND pair masks the slower input. However, if two signals arrive within a time interval shorter than the intrinsic feedback propagation delay of the NAND gate, the latch may enter a metastable state. A metastability filter is utilised downstream of the Set-Reset latch to mitigate this. Additionally, Table \ref{tab:wta_compare} presents a theoretical performance analysis of these implementations.

\begin{table}[h]
  \centering
  \footnotesize
  \caption{Theoretical analysis of WTA implementations}
  \label{tab:wta_compare}
  \resizebox{0.96\linewidth}{!}{
    \begin{tabular}{lccc}
      \toprule
      \textbf{Config.}
        & \shortstack{\textbf{Arbitration}\\\textbf{Depth}}
        & \textbf{Cell Count}
        & \shortstack{\textbf{Arbitration}\\\textbf{Latency}} \\
      \midrule
      TBA        & $\log_2 m$
                 & $m-1$
                 & $\log_2 m\,(\,d_{\text{Mutex}} + d_{\text{OR}} + d_{\text{C-Element}})$ \\
      Mesh-Like  & $m-1$
                 & $\dfrac{m(m-1)}{2}$
                 & $(m-1)\,d_{\text{Mutex}}$ \\
      \bottomrule
    \end{tabular}
  }
\end{table}

\subsubsection{\textbf{Four to two phase interface}}

The proposed classification functional module follows a four-phase handshake protocol. In this protocol, the activation and deactivation of the differential and SR race signals are each controlled by a Muller C-element. In contrast, the TM inference pipeline controller uses a two-phase handshake protocol; therefore, a TFF is integrated at the boundary to provide a reliable interface.

\begin{table}[ht]
  \centering
  \caption{Truth table of a two‐input Muller C‐element}
  \label{tab:2input_celement}
  \begin{tabular}{ccc}
    \toprule
    $a$ & $b$ & $c$ \\
    \midrule
    0 & 0 & 0 \\
    0 & 1 & $c_{\mathrm{prev}}$ \\
    1 & 0 & $c_{\mathrm{prev}}$ \\
    1 & 1 & 1 \\
    \bottomrule
  \end{tabular}
\end{table}

The Muller C-element, which is a state-holding component in asynchronous circuit design, exhibits the following characteristic behaviour: its output transitions to ‘1’ only when all inputs are ‘1’, drops to ‘0’ only when all inputs are ‘0’, and retains its previous state in all other scenarios. The truth table for a two-input Muller C-element is presented in Table~\ref{tab:2input_celement}.

\section{Experimental Results}

\subsection{Functional Verification}
\begin{figure}[h]
  \centering
  \subfloat[Multi-class TM]{%
    \includegraphics[width=\columnwidth]{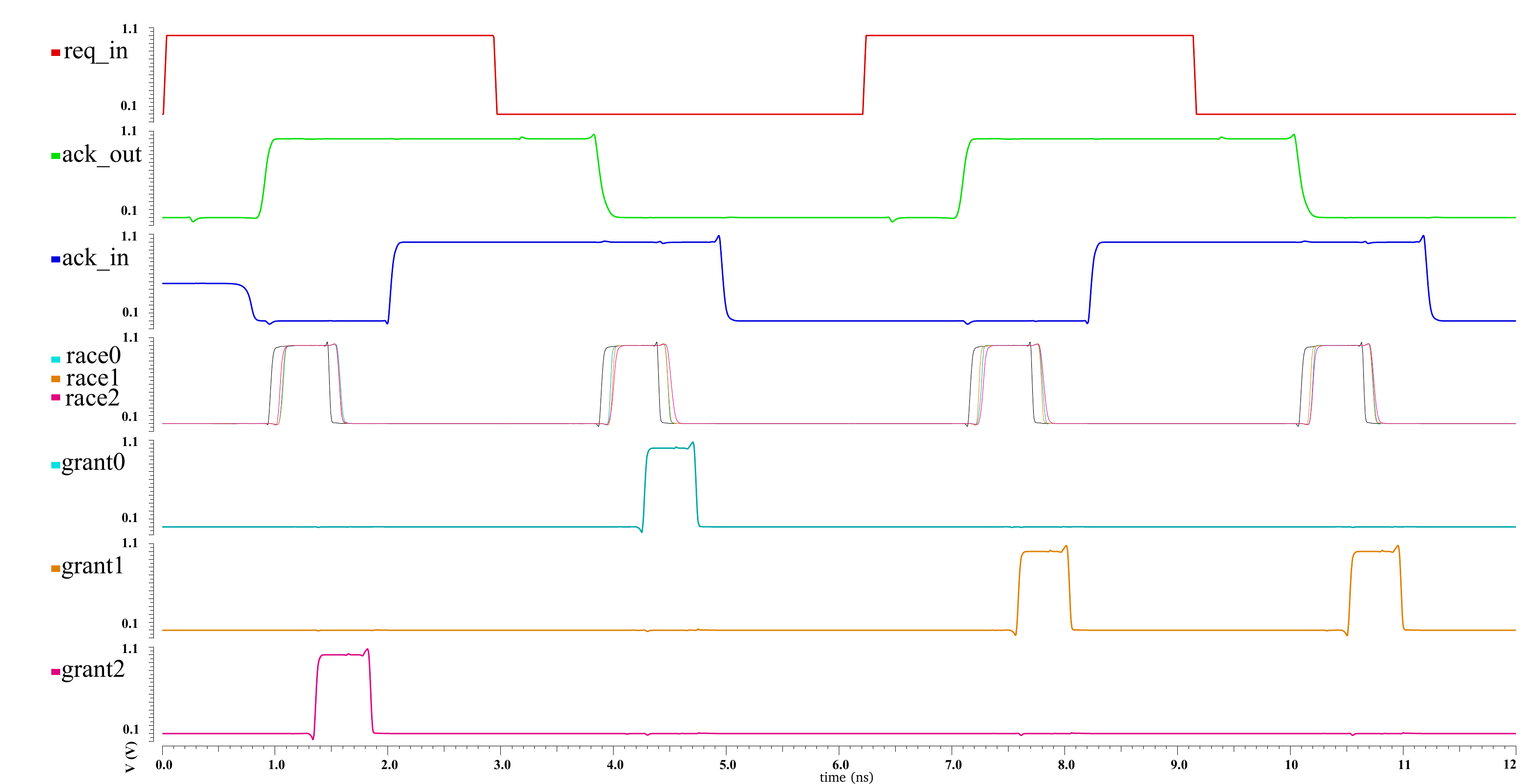}
    \label{fig:hdt_classic}
  }\\[1ex] 

  \subfloat[CoTM]{
    \includegraphics[width=\columnwidth]{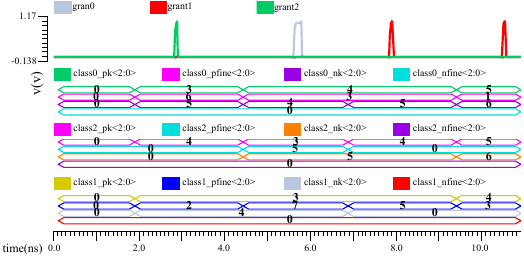}
    \label{fig:hdt_cotm}
  }
  \caption{Waveform of the proposed DT-domain architecture}
  \label{fig:timedomain}
\end{figure}

The verification circuits for the proposed architecture were built in Cadence Virtuoso using TSMC's 65nm CMOS technology and evaluated in an analogue mixed-signal simulation environment. The Iris classification dataset \cite{https://doi.org/10.1111/j.1469-1809.1936.tb02137.x} was adopted for functional verification, comprising 16 features, 12 clauses, and 3 classes. The resulting verification waveforms (see Fig. \ref{fig:timedomain}) illustrate two inference processes: (a) the multi-class TM, where class selection is determined through a Hamming-distance driven WTA arbitration mechanism (b) the CoTM, in which the class delay is first encoded in differential, then compressed using a LOD scheme into coarse and fine delay, and subsequently decided by WTA arbitration. For the identical input feature vector, both TM algorithms and their hardware realisations performed the correct prediction, supporting the target class sequence (2, 0, 1, 1).

\begin{figure}[h]
  \centering
  \subfloat[Synchronous pipeline]{%
    \includegraphics[width=\columnwidth]{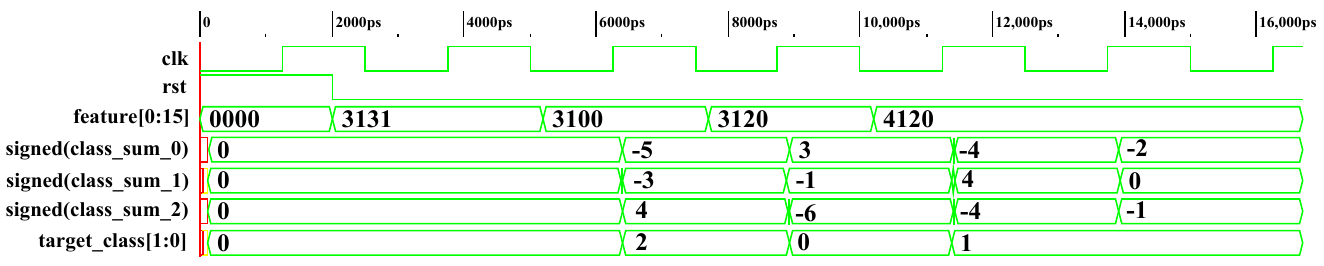}
    \label{fig:sync_classic}
  }\\[1ex] 

  \subfloat[Asynchronous BD pipeline]{
    \includegraphics[width=\columnwidth]{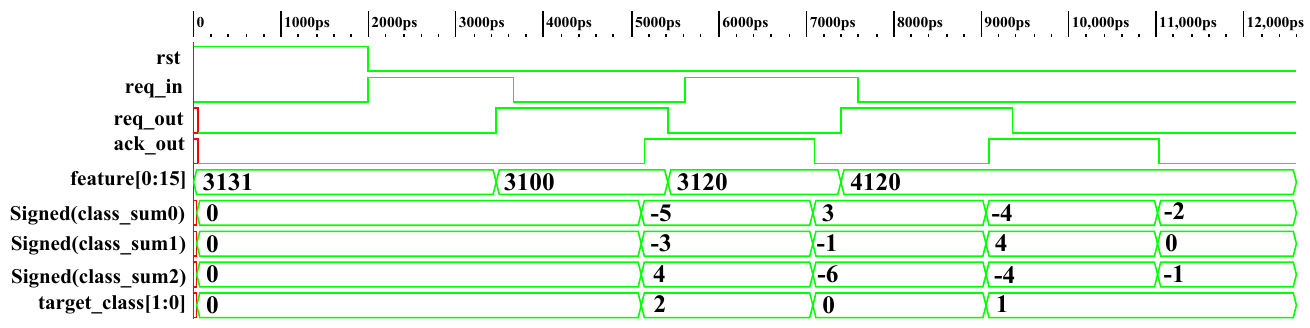}
    \label{fig:async_classic}
  }
  \caption{Waveform of digital implementation of multi-class TM}
  \label{fig:classicTM}
\end{figure}

\begin{figure}[h]
  \centering
  \subfloat[Synchronous pipeline]{%
    \includegraphics[width=\columnwidth]{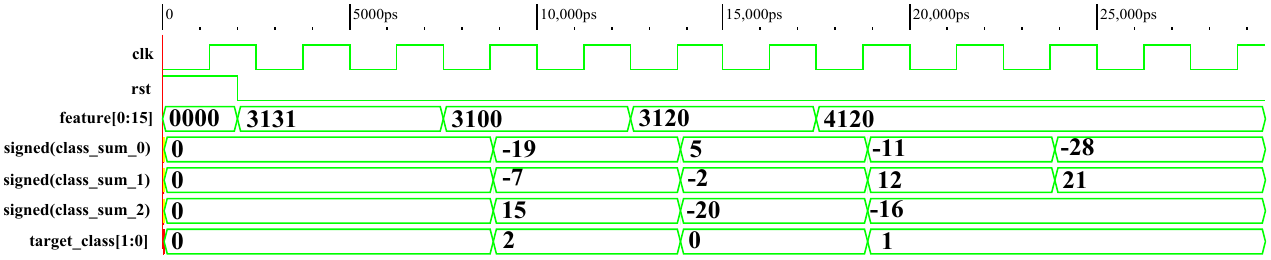}
    \label{fig:synccotm}
  }\\[1ex] 

  \subfloat[Asynchronous BD pipeline]{
    \includegraphics[width=\columnwidth]{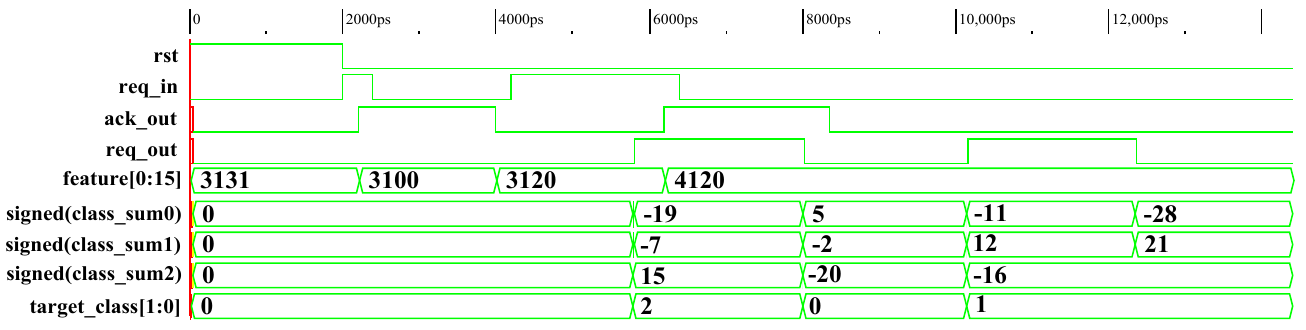}
    \label{fig:asynccotm}
  }
  \caption{Waveform of digital implementation of CoTM}
  \label{fig:Cotm}
\end{figure}

Purely digital-domain inference implementations for both multi-class TM and CoTM were realised as baselines. Functionally identical synchronous and asynchronous pipelines were developed in System Verilog, following Algorithm \ref{alg:click} \ref{alg:clauseE} \ref{alg:class}. Functional verification, logic synthesis, and physical implementation were executed with Cadence Xcelium, Genus, and Innovus. The synchronous and asynchronous digital-domain multi-class TM waveform is depicted in Fig.~\ref{fig:classicTM}, while the CoTM result is shown in Fig.~\ref{fig:Cotm}. The experiments confirm that all logically equivalent TM implementations achieve identical inference accuracy.

\subsection{Performance Evaluation}

\begin{table*}[!t]
\footnotesize        
\caption{Comparison With Some State-of-the-Art Works.}
\label{tab:architecture_comparison}
\centering
\resizebox{0.84\textwidth}{!}{
\begin{tabular}{@{}lcccccc@{}}  
\toprule
\textbf{Parameter} &
\textbf{\cite{9201533}} &
\textbf{\cite{10239577}} &
\textbf{\cite{10318175}} &
\textbf{\cite{9474126}} &
\textbf{Proposed} &
\textbf{Proposed} \\ \midrule
Architecture          & Async QDI & Async BD & Sync & Async QDI & Async BD & Async BD \\
Computing Domain      & Digital   & Digital  & Time & Digital   & Time     & Hybrid \\
Technology (nm)       & 65        & 28       & 65   & 65        & 65       & 65 \\
Voltage (V)           & 1.2       & 0.9      & 1.2  & 1.2       & 1.0      & 1.0 \\
Energy Eff. (TOP/J)   & 1.87      & 0.42     & 116  & 873       & 3329     & 750.79 \\
ML Algorithm          & CNN       & SNN      & BNN  & Multi-class TM & Multi-class TM & CoTM \\ \bottomrule
\end{tabular}}
\end{table*}

The inference process of TMs are evaluated as:
\begin{equation}
  \operatorname{Throughput}_{\mathrm{TM}}
  = 2\,F\,C\,K\,f_{\text{infer}}
  \label{eq:throughput}
\end{equation}
where \(F\) represents the number of input features, \(C\) is the number of clauses, \(K\) is the number of classes, and \(f_{\text{infer}}\) is the effective inference frequency.
\begin{equation}
  \operatorname{EnergyEfficiency}_{\mathrm{TM}}
  = \frac{\operatorname{Throughput}_{\mathrm{TM}}}{1000\,P}
  \label{eq:energy-eff}
\end{equation}
Where \(P\) is the average power consumption. Throughput is measured in giga operations per second (GOp/s), and energy efficiency is expressed in tera operations per joule (TOp/J).

Table~\ref{tab:tm_perf} summarises the performance metrics of various
implementations, covering fully digital synchronous and asynchronous BD baselines, as well as the proposed architectures for both multi-class TM and CoTM.

\begin{table}[h]
  \centering
  \caption{Performance summary}
  \label{tab:tm_perf}
  \resizebox{0.96\linewidth}{!}{%
  \begin{tabular}{@{}lcc@{}}
    \toprule
    \textbf{Implementation} & \textbf{Throughput} GOp/s & \textbf{Energy Efficiency} TOp/J \\
    \midrule
    Multi-class, synchronous           & 380   &  948.61 \\
    Multi-class, asynchronous BD       & 510   & 1381.65 \\
    Multi-class, proposed              & 402   & 3290.00 \\
    \midrule
    CoTM, synchronous                  & 230   &  304.65 \\
    CoTM, asynchronous BD              & 350   &  397.60 \\
    CoTM, proposed                     & 419   &  750.79 \\
    \bottomrule
  \end{tabular}}% end resizebox
\end{table}

Under identical functionality, the proposed architecture delivers substantial energy efficiency while sustaining or enhancing inference throughput. For multi-class TM, energy efficiency rises by 247\% over the synchronous digital baseline, with a throughput increase of 5.8\%. Compared to the asynchronous BD architecture, the proposed design sacrifices a 21\% throughput, improving energy efficiency by 138\%. In CoTM, the architecture simultaneously boosts throughput by 82\% and energy efficiency by 146\% versus the synchronous reference. Compared to the asynchronous BD counterpart, this approach improves 20\%  throughput and 89\% energy efficiency. Therefore, across both TM variants, this approach almost matches or exceeds the digital alternatives all around.

\subsection{Stat-of-the-art Work Comparison}

Table~\ref{tab:architecture_comparison} compares the proposed designs with several state-of-the-art ML accelerators. \cite{9201533} removes the global clock in a CNN engine by introducing a Dual-Rail data path for the MAC array. \cite{10239577} adopts an asynchronous pipeline for implementing SNN, where click-element-based pipelines facilitate an event-driven communication mechanism between neuron layers. Although both works benefit from clock removal and leverage computational sparsity, their arithmetic operations—matrix MACs in \cite{9201533} and membrane-potential updates in \cite{10239577}—are still confined to the digital domain, which limits their further energy efficiency.

In contrast, \cite{10318175} presents a time-domain accelerator for BNNs. This approach employs delay-encoded weights through a laddered inverter network, followed by delay accumulation and a WTA arbitration scheme. This time-encoded processing achieves a two-order-of-magnitude improvement in energy efficiency compared to other digital-domain NN architectures.

The \cite{9474126} is an ultra-low-power, multi-class TM edge chip using an asynchronous Dual-rail data path. Due to the lightweight nature of TM inference and the asynchronous characteristics, the design outperforms traditional NN accelerators from the aspect of energy consumption.

The final two columns of the table present the proposed architectures. The multi-Class TM design requires only Hamming-distance evaluation and arbitration. Integrating the time-domain logic in \cite{10666312} with an asynchronous BD controller achieves fully time-domain computation, achieving the highest energy efficiency, several orders of magnitude greater than all other surveyed architectures. The CoTM version employs a hybrid digital-time-domain approach, resulting in significant energy efficiency and throughput improvements compared to functionally equivalent pure-digital synchronous and asynchronous implementations.

\section{Conclusion}

This work introduces effective approaches to mitigate the energy overhead associated with the extensive arithmetic operations required in TM classification. The design illustrates that a fully time-domain computing flow achieves optimal energy efficiency for multi-class TM through Hamming-distance evaluation and arbitration. Conversely, for CoTM, which relies on MAC operations over a signed weight matrix, the proposed design employs a hybrid digital-time-domain approach. This approach processes the sign and magnitude components digitally, followed by differential encoding and LOD compression of the delay paths. A WTA arbitration system manages the final decision-making in the time domain.

To provide a comprehensive comparison, the analogous functionality of each TM algorithm was implemented digitally in both synchronous and asynchronous styles. This demonstrates the advantages of the proposed designs while offering additional options for designers who prefer digital-domain computing.

The aforementioned methodology confirms that asynchronous control combined with time-domain computation makes TM inference well-suited for resource-constrained edge platforms. This study outlines a clear path toward scalable, low-power ML deployments at the edge, bridging the gap between conventional ML accelerators and emerging event-driven time-domain hardware.

\bibliographystyle{IEEEtran}
\bibliography{ref.bib}

\end{document}